\pdfoutput=1
\documentclass[letterpaper]{article}
\usepackage{aaai2027}
\nocopyright
\usepackage[hyphens]{url}
\usepackage{graphicx}
\usepackage{natbib}
\usepackage{caption}
\usepackage{algorithm}
\usepackage{algorithmic}
\usepackage{amsmath}
\usepackage{amssymb}
\usepackage{booktabs}
\usepackage{multirow}
\usepackage{tikz}
\usepackage{placeins}
\usepackage{tabularx}
\usetikzlibrary{positioning,arrows.meta,shapes,shapes.geometric,fit,backgrounds,calc}
\makeatletter%
 
\begin{document}
\title{Fix, Not Just Resample: Bounded Compiler-Guided Repair in Lean 4}
\author{
    Haotian Deng\textsuperscript{\rm 1},
    Chris Farber\textsuperscript{\rm 1},
    Sonal Narasimha\textsuperscript{\rm 1},
    Samskrithi Sivakumar\textsuperscript{\rm 1},
    Yexiang Xue\textsuperscript{\rm 1}
}
\affiliations{
    \textsuperscript{\rm 1}Purdue University\\
    \{deng254, cfarber, narasim9, sivaku17, yexiang\}@purdue.edu
}
\maketitle

\begin{abstract} Automated theorem provers often rely on sampling language
models, but proof assistants like Lean reject each failed step with a precise
error message --- information that pure sampling discards. To overcome this,
we introduce a bounded, online step-level repair module. Our search samples
$K$ candidate tactics at each step and repairs a failed tactic at most $R$
times. Instead of discarding failed tactics, we feed the compiler's error
message to a pretrained repair model to fix individual rejected steps in
place. This approach improves the solve rate over a sampling-only baseline
across all tested candidate budgets ($K \in \{1,3,5,8\}$), achieving up to a
$32\%$ relative gain and solving $82$ additional theorems on a
$1{,}990$-theorem subset of the LeanDojo benchmark. Repair is not free: it
trades extra runtime for reachable proofs. But the trade is predictable:
repair pays most when sampling is constrained; its return is front-loaded,
with the first repair attempt capturing $84\text{--}96\%$ of the total gain;
and the gain comes almost entirely from shallow tactic-execution failures
rather than structural errors such as type mismatches. This targeted repair
successfully replicates on a second benchmark (Lean Workbook), nearly
doubling the solve rate with zero regressions.
\end{abstract}

\section{Introduction}

A growing share of AI systems produce output that a machine can check for
correctness automatically: generated code can be run against a test suite
\cite{chen_evaluating_2021}, and a generated math proof can be checked step by
step by a proof assistant. Because checking is cheap and automatic, the standard
way to solve a hard problem with such a system is simply to have the model
guess many times and keep whichever guess the checker accepts. This works, but
it runs into a wall: every guess is drawn from the same fixed model, so guesses
cluster around what that model already tends to produce, and solutions that lie
outside its distribution are hard to discover no matter how many times you
sample \cite{brown_large_2024}. Within any realistic budget, many problems the
model could in principle solve
go unsolved, because a correct candidate is never drawn.

We study this problem in neural theorem proving, where an AI model tries to
prove a formal math statement using the Lean proof assistant. A typical proof
in Lean consists of \emph{tactics} --- single proof steps, similar to lines of
code. The Lean compiler steps through the proof one tactic at a time: for each
tactic the model proposes, it either accepts it, moving the proof forward, or
rejects it with a precise error message, much like a compiler error. Today's
neural provers in the ReProver family, built on
LeanDojo \cite{NeurIPS2023/YangAARPSSR23}, handle a rejected tactic exactly like
the guessing strategy above: they discard it and simply propose a new one. That
hits the same wall --- guessing more tactics has rapidly diminishing returns, so
many provable theorems are missed simply because the prover never happens to
guess the tactic that works.

A rejected tactic, however, is not a dead end: it comes back with Lean's error
message, which states exactly what went wrong. Instead of discarding that
message and guessing blindly again, we hand the failed tactic and its error
message to a second, already-trained model whose only job is to \emph{repair}
it --- to propose a fixed version of that one step. Repair is thus not a
cheaper substitute for sampling but a complement to it: a second,
error-conditioned source of candidates. This differs from prior
systems that also use compiler feedback. Those systems, detailed below, wait
until an \emph{entire} proof attempt has already failed, then run the whole
broken proof through a separate, heavyweight pipeline --- a multi-agent loop, a
symbolic rule engine, or extra training --- to patch it after the fact. We
instead repair a single broken step \emph{immediately, in the middle of the
search}, using an existing repair model with no extra training. This lets us
ask a precise \emph{allocation question} no prior work answers: the moment one
tactic fails, is it worth more to fix that tactic or to just try guessing
again?

Concretely: APOLLO \cite{ospanov_apollo_2025} takes a proof attempt that has
already failed end-to-end, breaks it into sub-lemmas with a multi-agent loop,
and re-solves the failing pieces with auxiliary solvers. PALM
\cite{lu_proof_2024} first generates a complete proof, then patches its
low-level errors afterward with symbolic, rule-based methods. APRIL
\cite{wang_learning_2026} trains a dedicated repair model on 260{,}000
supervised (failure, diagnostic, fix) tuples, but likewise applies it to whole
finished proofs, not to individual steps mid-search. In each case, existing
systems apply it to a whole failed proof after search has finished and embed
it in enough surrounding machinery that repair's own contribution is hard to
isolate. We reuse APRIL's pretrained repair model
exactly as released --- no extra training, no agent orchestration, no symbolic
solver --- and directly measure that allocation question.

We integrate a bounded repair loop into tree search, with two budgets: $K$, how
many fresh tactics are sampled at a state, and $R$, how many times a failed
tactic may be repaired before it is abandoned. Setting $R{=}0$ recovers the
pure-sampling baseline exactly, since no failed tactic is ever repaired. Our
findings are:

\begin{itemize}
    \item \textbf{Repair reliably expands what the prover can solve.} At every
    candidate budget $K$ we test, enabling repair recovers proofs that
    same-budget sampling does not reach, lifting \emph{Pass@1} --- the
    percentage of theorems proved within the search budget --- by 2.3--4.1
    percentage points, up to a 32\% relative gain at the tightest budget, and
    solving up to 82 additional theorems on our LeanDojo subset.
    \item \textbf{This coverage comes at a cost, measured fairly.} The repair
    model is slower than the tactic generator, so repair buys reachable
    proofs in exchange for wall-clock, and at large sampling budgets that same
    time could instead buy more samples. Comparing repair-on and repair-off at
    the same $K$ would hand repair extra compute for free, so we instead read
    every configuration against pure-sampling ones of similar wall-clock ---
    an \emph{iso-runtime} comparison.
    \item \textbf{The benefit is front-loaded and narrowly sourced.} A single
    repair attempt ($R{=}0\to1$) captures 84--96\% of everything repair ever
    provides, while further repairs saturate quickly; it is most valuable in
    the sample-constrained regime, where drawing more samples is not an
    option, and it comes almost entirely from a narrow, identifiable slice of
    failures --- shallow tactic-execution errors, not structural ones like
    type mismatches.
    \item \textbf{The effect generalizes.} A second, retrieval-free benchmark
    (Lean Workbook), with a weaker generator and an untuned budget, reproduces
    the same pattern, nearly doubling the solve rate with zero regressions,
    so the effect is not an artifact of one generator or search
    configuration.
\end{itemize}

Together these turn repair from an all-or-nothing feature into a cheap,
well-scoped policy: add one targeted repair step to a generation-first
prover, and lean on it hardest when the sampling budget is scarce.

\section{Background: Tactic-Based Theorem Proving}

Lean checks proofs with a small trusted kernel: a theorem is a typed statement
and a proof is a term of that type. Rather than write the term directly, users
work in \emph{tactic mode}, issuing tactics that transform the proof state (\S
Introduction) until all goals close. An invalid tactic is rejected with a
diagnostic---a type mismatch, unknown identifier, unsolved goal, or timeout.
The repair model consumes exactly this diagnostic. Neural provers turn the
interaction into search: at each state a model proposes candidate tactics and
search continues from accepted successors. Our question is how to split a fixed
runtime between sampling fresh tactics and repairing rejected ones.

\section{Related Work}
\subsection{Generation-First Provers and Retrieval}
LeanDojo \cite{NeurIPS2023/YangAARPSSR23} established the now-standard
retrieval-augmented Lean proving setup, and ReProver-style systems spend their
budget on sampling more tactics and filtering them with the compiler.
REAL-Prover \cite{shen_real-prover_2025} extends that pattern with a dedicated
retrieval stack for harder algebraic benchmarks. LeanNavigator
\cite{yin_generating_2025} pushes the same breadth-first logic at the data layer
by generating millions of synthetic theorems and proofs. CARTS
\cite{yang_carts_2025} addresses another breadth-side bottleneck by calibrating
tactic diversity and confidence so tree search does not waste budget on
redundant proof paths. These systems increase the number, diversity, or quality
of fresh attempts, but they do not refine failed tactics during search.

\subsection{Scaling via Reinforcement Learning and Heavy Compute}
Recent systems spend compute through larger search and training loops
instead: AlphaProof \cite{Nature/HubertRLMGEAJYHOTAJIBVLOPSCBADFYIMSASEEHNDP25}
combines large-scale reinforcement learning with test-time search;
InternLM2.5-StepProver \cite{wu_internlm25_2025} guides search with a critic
model and expert iteration; POETRY
\cite{Proc.38thInt.Conf.NeuralInf.Process.Syst./WangHZWYJZJJZ24} decomposes
proofs into sketch and subproof stages; ProofAug \cite{liu_proofaug_2025}
turns proposals into compatible semi-proofs for tree search; Aristotle
\cite{achim_aristotle_2025} and Seed-Prover \cite{chen_seedprover_2025} add
informal lemma generation, a geometry solver, or deep test-time reasoning; and
AxProverBase \cite{requena_minimal_2026} iteratively refines a minimal agent,
reporting diminishing returns as iterations increase. None of these isolate
whether proof repair is a better use of a fixed compute budget than
additional sampling; our study measures that allocation decision directly,
treating repair as a policy choice rather than a feature that is simply
always on.

\subsection{Compiler-Feedback Repair}
APOLLO \cite{ospanov_apollo_2025}, PALM \cite{lu_proof_2024}, and APRIL
\cite{wang_learning_2026} --- the closest repair systems, discussed above ---
all act on a whole failed proof outside the search, whereas we treat repair
as a bounded operator inside search. Lean Copilot \cite{song_lean_2025} integrates
language-model assistance into the interactive Lean workflow rather than
automated search. A complementary line prevents errors before they occur:
type-constrained decoding \cite{Proc.ACMProgram.Lang./MundlerJHKD25} and ChopChop
\cite{Proc.ACMProgram.Lang./NagyTN26a} constrain generation so invalid tactics
are never produced, whereas our setting repairs after the compiler rejects a
tactic.

\begin{figure*}[t]
\centering
\begin{tikzpicture}[
    font=\small,
    >=Stealth,
    queue/.style={
        cylinder, shape border rotate=90,
        aspect=0.25,
        draw=gray!60, fill=gray!10, thick,
        text width=2.4cm, align=center,
        minimum height=0.8cm, minimum width=2.4cm
    },
    core/.style={
        rectangle, rounded corners=4pt,
        draw=gray!60, fill=gray!10, thick,
        text width=2.0cm, align=center,
        minimum height=1.0cm
    },
    compiler/.style={
        diamond, aspect=1.6,
        draw=gray!60, fill=gray!10, thick,
        text width=1.6cm, align=center,
        inner sep=2pt
    },
    repair/.style={
        rectangle, rounded corners=4pt,
        draw=orange!70, fill=yellow!25, thick,
        text width=2.0cm, align=center,
        minimum height=1.0cm,
        font=\small\bfseries
    },
    budget/.style={
        diamond, aspect=1.6,
        draw=orange!80, fill=yellow!40, thick,
        text width=1.6cm, align=center,
        inner sep=2pt, font=\small\bfseries
    },
    queueout/.style={
        cylinder, shape border rotate=90,
        aspect=0.25,
        draw=gray!60, fill=gray!10, thick,
        text width=2.4cm, align=center,
        minimum height=0.8cm, minimum width=2.4cm
    },
    proved/.style={
        ellipse,
        draw=green!50!black, fill=green!15,
        line width=2pt,
        text width=1.6cm, align=center,
        minimum height=1.4cm,
        font=\small\bfseries, text=green!30!black
    },
    main/.style={->, thick, draw=gray!70},
    vanilla/.style={->, dashed, thick, draw=gray!50},
    repairarrow/.style={->, very thick, draw=orange!80},
    droparrow/.style={->, dashed, draw=gray!60},
    lbl/.style={font=\small, fill=white, inner sep=1pt},
]

\node[queue]    (Qin)    at (0,   0)     {Proof State\\Queue};
\node[core]     (Gen)    at (3.7,  0)    {Neural\\Tactic\\Generator};
\node[compiler] (Lean)   at (7.8,  0)    {Lean\\Compiler};
\node[queueout] (Qout)   at (12.6, -3)    {Proof State\\Queue};
\node[proved]   (Proved) at (12.6, 0)  {Proof\\Complete};
\node[budget]   (Budget) at (7.8, -3)  {Check\\Repair\\Budget};
\node[repair]   (Diag)   at (3.8, -3)  {Repair\\Model};

\begin{scope}[on background layer]
    \node[
        draw=orange!60, fill=yellow!8, thick,
        rounded corners=8pt,
        inner sep=8pt,
        fit=(Budget)(Diag),
        label={[font=\small\bfseries, text=orange!70!black]below:%
            Iterative Tactic Repair Module~---~\textit{Our Contribution}}
    ] (RepairBox) {};
\end{scope}

\draw[main] (Qin.east) -- (Gen.west)
    node[midway, lbl, above=4pt] {State};

\draw[main] (Gen.east) -- (Lean.west)
    node[midway, lbl, above=13pt, xshift=11pt] {Candidate Tactic};

\draw[main, draw=green!60!black] (Lean.south east) -- ++(2,0) 
    node[pos=.75, lbl, yshift=-8pt]
        {\textcolor{green!60!black}{Valid (push new state)}}
    -- (Qout.north);

\draw[main] (Lean.east) -- (Proved.west)
    node[midway, lbl, above=4pt] {Goal Solved};

\draw[repairarrow] (Lean.south) -- (Budget.north)
    node[pos=0.35, lbl, left=1pt]
        {\textcolor{orange!80!black}{Error (ours: repair)}};

\draw[droparrow, draw=red!70!black] (Lean.south) -- ($(Lean.south)+(0,-0.48)$) -- ++(2, 0) 
    node[pos=.85, lbl, yshift=6pt]
        {\textcolor{red!70!black}{Error (baseline: drop)}}
    -- (Qout.north west);

\draw[repairarrow] (Budget.west) -- (Diag.east)
    node[midway, font=\small, below=14pt] {State + Tactic + Error};

\draw[repairarrow] (Diag.west) -- ++(-1.0,0) -- ++(0, 1.7) -- ++(3, 0)
    node[midway, lbl, yshift=8pt] {Fixed Tactic} -- (Lean.south west);

\draw[droparrow, draw=gray!60] (Budget.east) -- (Qout.west)
node[midway, font=\small, anchor=center, xshift=-2pt, yshift=-20pt, fill=white, inner sep=1pt] {{Budget Exceeded}};

\draw[vanilla] (Qout.south) -- ++(0, -1.3)
    -- ++(-12.6, 0) -- (Qin.south)
    node[pos=0.5, lbl, below=4pt] {Next Iteration};

\end{tikzpicture}
\caption{Architecture of the bounded iterative repair system. Solid arrows show
the main proof-search pipeline; the dashed baseline path shows tactics dropped
without repair; bold orange arrows highlight the novel repair module.
$K$ tactics are sampled per state; each failed tactic triggers at most $R$
repair attempts before being abandoned.}
\label{fig:architecture}
\end{figure*}
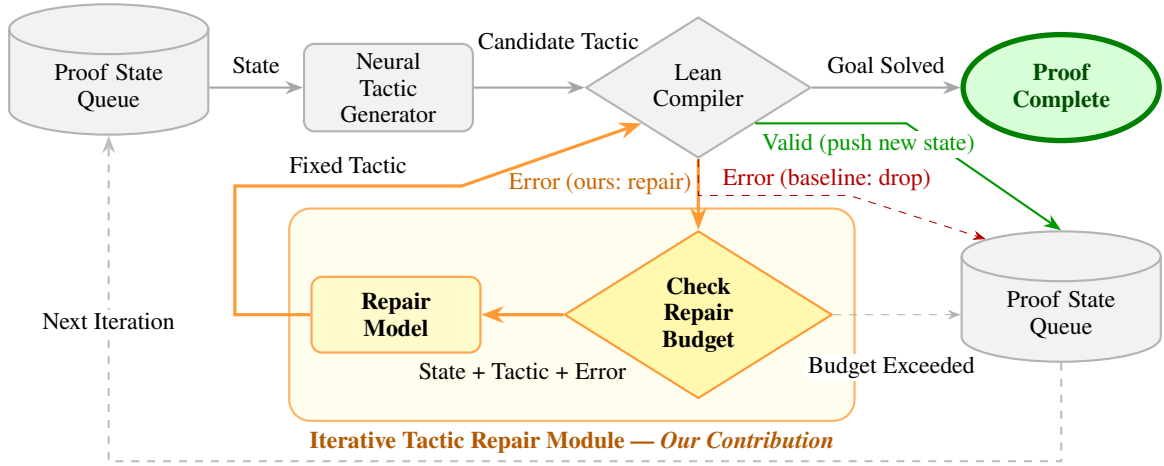 
\section{Methodology}
We define a bounded iterative repair loop integrated into a priority-queue search
algorithm.
\subsection{Baseline Search Architecture}
The baseline prover maintains a priority queue of proof states. At each step, it:
\begin{enumerate}
    \item Dequeues the state $s$ with the highest priority.
    \item Uses a premise retriever to find relevant lemmas $\mathcal{P}$.
    \item Queries a tactic generator to produce $K$ candidate tactics $\{t_1, \dots, t_K\}$.
    \item Evaluates each tactic in the Lean 4 environment.
\end{enumerate}

\subsection{Iterative Repair Module}
When a tactic $t_i$ fails, the Lean compiler returns an error message $e_i$. Our repair module $M_R$ takes the tuple $(s, t_i, e_i)$ and generates a corrected tactic $t'_i$.

\subsubsection{Formal Parameters: K and R}
To prevent the search from being dominated by repair attempts on a single unviable branch, we enforce two bounds:
\begin{itemize}
    \item \textbf{K (Candidate Budget):} The number of unique tactics initially sampled for each state.
    \item \textbf{R (Repair Budget):} The maximum number of times a specific tactic attempt can be iteratively repaired before it is abandoned.
\end{itemize}

Total tactic attempts per state are bounded by $K \times (1 + R)$. This bound
turns repair into an explicit depth budget rather than an unbounded retry loop.
When $R=0$, the system reduces to pure sampling. As $R$ increases, the search
spends more compute on refining existing failed tactics and less of a fixed
runtime budget on exploring fresh samples.

\subsubsection{The Repair Loop}
The full algorithm is described in Algorithm~1, and Figure~\ref{fig:architecture}
illustrates the overall system architecture.

\begin{algorithm}[h]
\caption{Priority-Queue Search with Bounded Iterative Repair}
\begin{algorithmic}[1]
\STATE $Q \gets \{s_{init}\}$
\WHILE{$Q \neq \emptyset$ and not timeout}
    \STATE $s \gets \text{pop\_best}(Q)$
    \STATE $\text{Candidates} \gets \text{SampleTactics}(s, K)$
    \FORALL{$t \in \text{Candidates}$}
        \STATE $r \gets 0$
        \WHILE{$r \le R$}
            \STATE $(\text{success}, \text{result}, e) \gets \text{LeanEval}(s, t)$
            \IF{success}
                \STATE \textbf{if} result is solved \textbf{return} proof
                \STATE $Q \gets Q \cup \text{result}$
                \STATE \textbf{break}
            \ELSE
                \IF{$r = R$}
                    \STATE \textbf{break}
                \ENDIF
                \STATE $t \gets \text{Repair}(s, t, e)$
                \STATE $r \gets r + 1$
            \ENDIF
        \ENDWHILE
    \ENDFOR
\ENDWHILE
\end{algorithmic}
\end{algorithm}

\subsection{The Repair Prompt}
When a tactic fails, the repair model $M_R$ is queried with a fixed chat
template that supplies exactly three pieces of information from the failure: the
Lean proof state (infoview) at the failing position, the failed tactic itself,
and Lean's verbatim error message. Because the repair model is a supervised,
instruction-tuned model \cite{wang_learning_2026}, we query it through the same
chat template it was fine-tuned on---a system role and a user role holding the
three slots above---rather than a bare feature string; the roles are the model's
trained interface, not optional scaffolding. It returns a corrected proof,
sometimes prefixed by a short natural-language explanation; we keep only the
final \texttt{lean} block and execute it \emph{as a tactic from the failing proof
state}, not as a replacement for the whole proof, so repair remains a per-node,
in-search operator (Algorithm~1). If the extracted block is not a valid tactic
sequence for that state, Lean rejects it and the repair simply fails.
Figure~\ref{fig:repair-prompt} shows the template with a minimal example
substituted into the slots.

\begin{figure}[tb]
\centering
\fbox{\begin{minipage}{0.94\columnwidth}
\ttfamily\small
\textbf{[system]} You are a Lean 4 programmer diagnosing a single failing
proof. Assume you only see the incorrect proof text, the infoview state near
the failure, and Lean's error message.\\[3pt]
\textbf{[user]} Provide the full corrected Lean 4 theorem/proof in a single
\texttt{lean} code block.\\[2pt]
Incorrect proof: \textit{rw [add\_comm]}\\
Infoview state: \textit{n : Nat $\vdash$ n + 0 = n}\\
Line at error: \textit{rw [add\_comm]}\\
Lean error: \textit{tactic 'rewrite' failed, motive is not type correct \ldots}\\[3pt]
\textbf{[model output]} \texttt{```lean}\ \textit{rw [Nat.add\_zero]}\ \texttt{```}
\end{minipage}}
\caption{The repair prompt actually sent to $M_R$: a system role, a user role
with the three failure slots (proof state, failed tactic, Lean error), and the
model's returned correction. Italicized text marks the per-failure values; here
a minimal example is substituted.}
\label{fig:repair-prompt}
\end{figure}

\section{Experimental Setup}
Our experimental design aims to quantify the search-time impact of iterative tactic repair across different branching factors ($K$) and repair depths ($R$).

\subsection{Benchmarks}
We evaluate our system on the LeanDojo Benchmark
\cite{NeurIPS2023/YangAARPSSR23}, specifically a randomly sampled subset of
2,000 theorems. This benchmark provides a diverse set of formal mathematical
statements and their ground-truth proofs in Lean 4. The subset was sampled to
avoid data leakage with the repair model's training set.

To test whether the repair effect is specific to this one benchmark, we also
evaluate on a second, harder benchmark: Lean Workbook \cite{ying_lean_2024}, a
corpus of natural-language competition-style math problems formalized into
Lean~4. We generate a 98-theorem validation split and an 89-theorem held-out
test split (187 of 200 sampled dataset rows compile as valid Lean~4
statements; the rest are dropped) using a separate, retrieval-free generator
configuration described next.

\subsection{Model Configurations}
We use the public ReProver \texttt{byt5-small} tactic generator and its premise
retriever \cite{NeurIPS2023/YangAARPSSR23} as the baseline prover. For the repair
module, we employ the \texttt{gAPRIL-wo-exp} model \cite{wang_learning_2026}, a
4-billion parameter model trained on supervised proof-repair traces. We do not
train or fine-tune either model; the training-free claim refers to integrating a
pretrained repair module into search.

\paragraph{Implementation and determinism.}
The generator decodes with beam search (beam width and number of returned
tactics both equal to $K$); the repair model decodes greedily. Both are
\emph{deterministic}---no sampling and no random seeds---so the entire search is
reproducible and each reported number comes from a single run that reproduces
exactly (there is no run-to-run variance to average over). Input/output lengths
are capped at 2048/512 tokens. All source code (the prover, the repair
integration, and the analysis scripts that produce every table and figure) and
the run manifests will be released, with library versions pinned.

We perform a grid sweep over the following parameters:
\begin{itemize}
    \item \textbf{K (Candidates):} $\{1, 3, 5, 8\}$
    \item \textbf{R (Repairs):} $\{0, 1, 2, 3\}$
\end{itemize}
The $R=0$ configuration serves as the vanilla ReProver baseline for each $K$.
In the figures and run labels, a configuration with $K$ candidates is written
\texttt{tac}$K$ (e.g.\ \texttt{tac5} denotes $K{=}5$).

\paragraph{Lean Workbook configuration.}
For the Lean Workbook generalization experiment (\S\ref{sec:lean-workbook}) we pair
the same \texttt{gAPRIL-wo-exp} repair model with a deliberately harder
generator configuration: the same priority-queue (best-first) search, but with
the premise retriever removed and a single fixed candidate budget $K{=}5$,
swept over $R\in\{0,1,2\}$. Removing retrieval makes the generator strictly
weaker than the LeanDojo configuration above, so this run tests repair against
a lower, retrieval-free baseline rather than reusing the retrieval-augmented
setup.

\paragraph{Common theorem set.}
A small number of theorems are discarded by at least one $K \times R$
configuration (e.g.\ tracing failures or non-theorem statements), so every
table reports the common set of theorems present in every configuration's run
export, rather than the full 2,000 sampled theorems, so that all
configurations are scored on identical problems.

\subsection{Metrics}
\begin{figure}[t]
    \centering
    \resizebox{\linewidth}{!}{%
\begin{tikzpicture}[
  >=Stealth,
  proofstate/.style={draw=blue!60!black, circle, fill=blue!20, minimum size=5pt, inner sep=0pt},
  errorstate/.style={draw=red!70!black, diamond, fill=red!35, minimum size=6pt, inner sep=0pt},
  successstate/.style={draw=green!55!black, star, star points=5, star point ratio=2.3, fill=green!65, minimum size=9pt, inner sep=0pt},
  genedge/.style={->, semithick},
  repairedge/.style={->, dashed, semithick, blue!70!black},
]
\node[font=\small, anchor=north] at (0.425,-1.540) {(a) $R{=}0$};
\node[font=\small, anchor=north] at (3.145,-1.540) {(b) $R{=}1$};
\node[proofstate] (L1_0) at (0.000,0.000) {};
\node[errorstate] (L1_ERROR_0) at (0.850,0.000) {};
\node[errorstate] (L1_ERROR_1) at (0.850,-0.420) {};
\node[errorstate] (L1_ERROR_2) at (0.850,-0.840) {};
\node[proofstate] (L2_0) at (1.870,0.000) {};
\node[proofstate] (L2_1) at (3.570,0.000) {};
\node[proofstate] (L2_2) at (4.420,0.000) {};
\node[errorstate] (L2_ERROR_0) at (2.720,0.000) {};
\node[errorstate] (L2_ERROR_1) at (2.720,-0.420) {};
\node[errorstate] (L2_ERROR_3) at (2.720,-0.840) {};
\node[errorstate] (L2_ERROR_4) at (3.570,-0.420) {};
\node[successstate] (L2_FINISH_6) at (4.420,-0.420) {};
\draw[genedge] (L1_0.east) -- (L1_ERROR_0.west);
\draw[genedge] (L1_0.east) -- (L1_ERROR_1.west);
\draw[genedge] (L1_0.east) -- (L1_ERROR_2.west);
\draw[genedge] (L2_0.east) -- (L2_ERROR_0.west);
\draw[genedge] (L2_0.east) -- (L2_ERROR_1.west);
\draw[repairedge] (L2_ERROR_1.east) -- (L2_1.west);
\draw[genedge] (L2_0.east) -- (L2_ERROR_3.west);
\draw[repairedge] (L2_ERROR_3.east) -- (L2_ERROR_4.west);
\draw[genedge] (L2_1.east) -- (L2_2.west);
\draw[genedge] (L2_1.east) -- (L2_FINISH_6.west);
\node[proofstate] (L1_0) at (0.000,0.000) {};
\node[errorstate] (L1_ERROR_0) at (0.850,0.000) {};
\node[errorstate] (L1_ERROR_1) at (0.850,-0.420) {};
\node[errorstate] (L1_ERROR_2) at (0.850,-0.840) {};
\node[proofstate] (L2_0) at (1.870,0.000) {};
\node[proofstate] (L2_1) at (3.570,0.000) {};
\node[proofstate] (L2_2) at (4.420,0.000) {};
\node[errorstate] (L2_ERROR_0) at (2.720,0.000) {};
\node[errorstate] (L2_ERROR_1) at (2.720,-0.420) {};
\node[errorstate] (L2_ERROR_3) at (2.720,-0.840) {};
\node[errorstate] (L2_ERROR_4) at (3.570,-0.420) {};
\node[successstate] (L2_FINISH_6) at (4.420,-0.420) {};
\begin{scope}[shift={(4.870,0.000)}]
\node[font=\small, anchor=west] at (0,0) {Legend};
\draw[genedge] (0,-0.25) -- (0.35,-0.25);
\node[font=\small, anchor=west] at (0.45,-0.25) {Generator};
\draw[repairedge] (0,-0.5) -- (0.35,-0.5);
\node[font=\small, anchor=west] at (0.45,-0.5) {Repair};
\node[proofstate] at (0.18,-0.78) {};
\node[font=\small, anchor=west] at (0.45,-0.78) {Proof state};
\node[successstate] at (0.18,-1.05) {};
\node[font=\small, anchor=west] at (0.45,-1.05) {Solved};
\node[errorstate] at (0.18,-1.32) {};
\node[font=\small, anchor=west] at (0.45,-1.32) {Failed};
\end{scope}
\end{tikzpicture}
}
    \caption{Visualization of search-tree traces for \texttt{Set.not\_subsingleton\_iff}
    at $K{=}3$, from the raw run logs. Nodes are proof states, encoded by shape
    and color: blue circles are open states, red diamonds are failed states, and
    the green star is the solved state. Solid black edges are generator tactics;
    dashed blue edges are repair attempts. Without repair~(a) every path
    dead-ends in a failed state; a single repair attempt~(b) recovers a
    failed tactic and reaches the solved state.}
    \label{fig:repair-tree-example}
\end{figure}
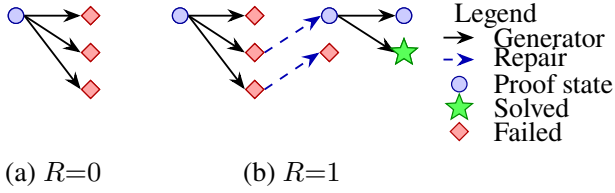

We evaluate performance along four axes:
\begin{enumerate}
    \item \textbf{Pass@1:} the percentage of theorems successfully proved
    within the search budget, reported for every $K \times R$ configuration in
    Table~\ref{tab:main_results} and, on a second benchmark, in
    Table~\ref{tab:lean_workbook}.
    \item \textbf{Measured mean wall-clock per theorem:} the primary cost
    axis. Because repair spends extra time, comparing repair-on and
    repair-off at the same $K$ is unfair; we instead read configurations
    against pure-sampling ones of similar wall-clock (an \emph{iso-runtime},
    or equal-runtime, comparison). This is reported as a single number per
    configuration in Table~\ref{tab:main_results}'s $T(s)$ column and, as a
    full budget-vs-coverage curve, in Figure~\ref{fig:coverage-time}.
    \item \textbf{Net Delta:} the number of theorems gained by repair
    relative to the no-repair baseline at the same $K$, minus the baseline
    solves lost by that repair configuration. Reported cumulatively (from
    $R{=}0$) in Table~\ref{tab:main_results} and Figure~\ref{fig:repair-saturation},
    and broken down step by step (the marginal gain of each individual
    repair attempt) in Table~\ref{tab:repair_dynamics}.
    \item \textbf{Generalization:} whether the repair effect holds on an
    independent, out-of-distribution benchmark, rather than being an artifact
    of one benchmark/generator/search combination. Tested by repeating the
    same $K$-vs-$R$ comparison on Lean Workbook with a different,
    retrieval-free generator (\S\ref{sec:lean-workbook},
    Table~\ref{tab:lean_workbook}).
\end{enumerate}
Beyond these quantitative metrics, we complement the results with two
qualitative analyses: a breakdown of repair's fix rate by failure category
(Table~\ref{tab:failure_taxonomy}), showing which kinds of rejected tactics
repair actually recovers, and a search-tree visualization of a single repair
recovering a proof that pure sampling misses
(Figure~\ref{fig:repair-tree-example}).

\subsection{Hardware and Budget}
All experiments run on a single \textbf{NVIDIA Tesla V100-PCIE (32\,GB)} GPU
(CUDA 12.6, driver 575.57) hosted on a 24-core Intel Xeon Gold 6126 with
754\,GB RAM, under Rocky Linux 9.6. The software stack is Python 3.11,
PyTorch 2.9.1, HuggingFace Transformers 4.56, and LeanDojo driving Lean~4; all
package versions are pinned in the released code. We enforce
a wall-clock timeout of 600 seconds per theorem to ensure a fair comparison that
accounts for the computational cost of the repair module.

\subsection{External Validity}
The exact fix rates and crossover points depend on the benchmark, tactic
generator, repair model, and hardware, so we do not present the LeanDojo sweep
as a universal constant. A second benchmark with a different generator and a
fixed, untuned budget reproduces the same qualitative pattern
(\S\ref{sec:lean-workbook}), so the transferable claim is the recipe itself: a
lightweight in-search repair step improves on the generation-first baseline,
its benefit is front-loaded in the first attempt, and the same iso-runtime
analysis lets any prover combining stochastic tactic generation with compiler
feedback decide when repair is worth invoking.

\section{Results}
\label{sec:results}
The central result is that a bounded in-search repair step improves on the
generation-first baseline at every budget. In Table~\ref{tab:main_results} the
$R{=}0$ column is a pure-sampling, LeanDojo/ReProver-style prover, and enabling
repair lifts Pass@1 above it for all four candidate budgets -- by 2.3--4.1 percentage points
and 45--82 net theorems on the shared 1{,}990-theorem set (the Fix and Net
columns count the proofs repair uniquely unlocks). The gain is also cheap,
because it is front-loaded: the first repair attempt ($R{=}0\to1$) captures the
large majority of the benefit repair ever provides, and every attempt after it
buys only marginal further solves at rising wall-clock cost. We read each cell by
its measured wall-clock so repair is compared against pure sampling of similar
cost, not given extra compute for free.

Every cell is computed over a single common set of 1{,}990 theorems attempted
under all $K \times R$ configurations, so all rates are directly comparable
across budgets. The Net Delta column exposes unique repair-only unlocks even
when average pass rates are close, and the rare negative regime where repair
wastes runtime on failed refinements better spent sampling. Because decoding is
deterministic, the per-theorem outcomes are paired and allow an exact
significance test: a McNemar test on the Fix/Brk discordant pairs rejects the
no-effect null at every budget ($R{=}1$ vs.\ $R{=}0$; $p<10^{-11}$ for all
$K$).

\begin{table}[tbp]
\centering
\setlength{\tabcolsep}{3pt}
{\small
\begin{tabular}{@{}cc r r c r r r@{}}
\toprule
$K$ & $R$ & Atmp. & T(s) & P@1 & Fix & Brk & Net \\
\midrule
\multirow{4}{*}{1} & 0 & \multirow{4}{*}{1990} &  29.7 &  7.1 &  0 &   0 & \phantom{$+$}0 \\
                   & 1 &                        &  30.3 &  9.0 & 38 &   0 & $+$38 \\
                   & 2 &                        &  31.8 &  9.4 & 45 &   0 & $+$45 \\
                   & 3 &                        &  33.0 &  9.4 & 45 &   0 & $+$45 \\
\midrule
\multirow{4}{*}{3} & 0 & \multirow{4}{*}{1990} &  54.9 & 20.9 &  0 &   0 & \phantom{$+$}0 \\
                   & 1 &                        &  83.9 & 23.8 & 57 &   0 & $+$57 \\
                   & 2 &                        &  98.0 & 24.3 & 67 &   0 & $+$67 \\
                   & 3 &                        &  97.1 & 24.3 & 68 &   0 & $+$68 \\
\midrule
\multirow{4}{*}{5} & 0 & \multirow{4}{*}{1990} & 116.0 & 29.9 &  0 &   0 & \phantom{$+$}0 \\
                   & 1 &                        & 191.6 & 33.5 & 72 &   1 & $+$71 \\
                   & 2 &                        & 221.9 & 33.9 & 81 &   1 & $+$80 \\
                   & 3 &                        & 236.7 & 34.0 & 83 &   1 & $+$82 \\
\midrule
\multirow{4}{*}{8} & 0 & \multirow{4}{*}{1990} & 200.5 & 34.8 &  0 &   0 & \phantom{$+$}0 \\
                   & 1 &                        & 367.3 & 38.7 & 83 &   6 & $+$77 \\
                   & 2 &                        & 450.0 & 38.8 & 84 &   5 & $+$79 \\
                   & 3 &                        & 485.7 & 38.8 & 85 &   5 & $+$80 \\
\bottomrule
\end{tabular}
}
\caption{%
  LeanDojo results across all $K \times R$ configurations. The $R{=}0$ row of
  each $K$ block is the pure-sampling (LeanDojo/ReProver-style) baseline;
  repair ($R>0$) is measured against it.
  \textit{Atmp.}~=~theorems attempted;
  \textit{T(s)}~=~measured mean wall-clock per theorem (seconds);
  \textit{P@1}~=~Pass@1 (\%);
  \textit{Fix}/\textit{Brk}~=~theorems newly solved / regressed vs.\ the baseline;
  \textit{Net}~=~signed change in theorems solved.
  At every $K$, the $R{=}0{\to}1$ step accounts for 84--96\% of the total
  Net gain reached by $R{=}3$ (see below); \textit{Fix}/\textit{Net}
  are the most precise readout of this saturation where the rounded
  \textit{P@1} column loses resolution (e.g.\ $K{=}3$, $R{=}2$ vs.\ $R{=}3$).
}
\label{tab:main_results}
\end{table}

\begin{figure}[t]
    \centering
    \includegraphics{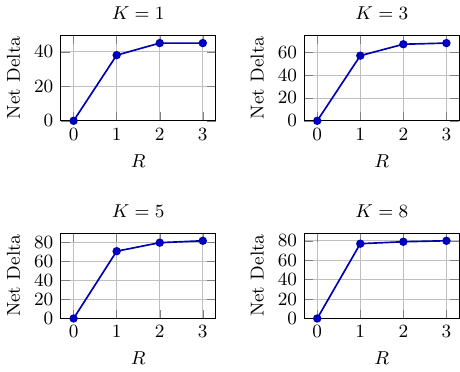}
    \caption{Net Delta versus repair depth $R$ for each candidate budget $K$.
    Every panel shows the same shape: a large gain from $R{=}0$ to $R{=}1$,
    followed by a flat or near-flat plateau.}
    \label{fig:repair-saturation}
\end{figure}

This improvement is also cheap, because it is front-loaded.
Figure~\ref{fig:repair-saturation} shows Net Delta rising sharply from $R{=}0$
to $R{=}1$ at every $K$, then flattening into a visible elbow, and
Table~\ref{tab:repair_dynamics} breaks the same pattern down step by step: the
first repair attempt alone accounts for the 84--96\% of the total $R{=}3$ gain
noted above. Under a fixed repair budget, spending it on a single repair attempt
is worth far more than spending it on a second or third, so essentially all of
the gain is available at $R{=}1$.

The elbow is sharpest at $K{=}8$, where the $R{=}1\to2$ step adds only 2.5\% of
the total three-repair gain, and weakest at $K{=}1$, where that same step still
adds 15.6\%. We report this rather than smooth it over: at the smallest,
most sample-constrained budget, a second repair attempt still recovers a small
but non-trivial number of additional proofs, even though the dominant effect
at every $K$ remains the jump from $R{=}0$ to $R{=}1$. The pattern is therefore
best read as ``diminishing,'' not ``vanishing,'' returns beyond the first repair.

\begin{figure}[t]
    \centering
    \includegraphics{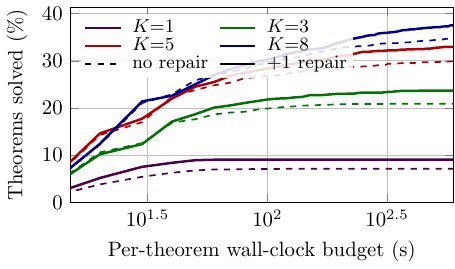}
    \caption{Iso-runtime coverage: theorems solved within a per-theorem
    wall-clock budget $T$ (log scale), for $K\in\{1,3,5,8\}$ with (solid) and
    without (dashed) one repair. Repair lifts coverage at matched wall-clock
    across the range at every $K$ shown, the gap widening as the budget grows
    (flat at $K{=}1$, whose short runtime saturates coverage early).}
    \label{fig:coverage-time}
\end{figure}

The $K \times R$ grid can also be read through a breadth-vs-depth lens: at a
fixed wall-clock budget, is the next unit of compute better spent on a repair or
on another sample? A fair comparison must cap each theorem's wall-clock equally,
since comparing configurations at their \emph{mean} runtime is misleading---the
mean is inflated by timed-out theorems. Figure~\ref{fig:coverage-time} plots
\emph{coverage}---the fraction of
theorems solved within a per-theorem budget $T$---for $K\in\{1,3,5,8\}$ with and
without repair. Under this matched-time comparison, adding one repair
improves coverage at essentially every budget for every $K$ shown, and the
margin widens with $T$ for the three largest budgets (e.g.\ at $K{=}8$, from
$+0.6$ points at $100$\,s to $+2.8$ at $600$\,s); the sole exception is a
tight ${\sim}50$\,s budget, where repair's per-state overhead delays some
solves past the deadline. At $K{=}1$ the gain is
present but flat rather than widening ($+2.0$ points at $100$\,s, $+1.9$ at
$600$\,s): mean per-theorem runtime at this budget is only tens of seconds
(Table~\ref{tab:main_results}), so coverage has already saturated well before
$100$\,s and the extra budget adds nothing further. Because the per-theorem
time budget is held
fixed, this gain reflects genuinely broader coverage, not the extra compute an
equal-$K$ comparison would give to repair. (Per-theorem wall-clock carries some
concurrency noise, so we read the curves as a trend.)

\subsection{Generalization to Lean Workbook}
\label{sec:lean-workbook}
The LeanDojo sweep uses one benchmark, one generator, and one search
configuration, so Table~\ref{tab:main_results} alone cannot rule out that the
repair effect is an artifact of that particular combination. Lean Workbook
tests this with a different benchmark (harder, natural-language-formalized
competition problems), a weaker generator (no premise retrieval), and a fixed,
untuned budget ($K{=}5$), evaluated on both a validation and a held-out test
split.

\begin{table}[tbp]
\centering
\setlength{\tabcolsep}{4pt}
{\small
\begin{tabular}{@{}l r r r r r@{}}
\toprule
Split & $R$ & Proved & P@1 & Fix & Brk \\
\midrule
\multirow{3}{*}{val (98)}  & 0 & 21 & 21.4\% & --  & -- \\
                           & 1 & 39 & 39.8\% & 18 & 0 \\
                           & 2 & 42 & 42.9\% & 21 & 0 \\
\midrule
\multirow{3}{*}{test (89)} & 0 & 17 & 19.1\% & --  & -- \\
                           & 1 & 27 & 30.3\% & 10 & 0 \\
                           & 2 & 28 & 31.5\% & 11 & 0 \\
\bottomrule
\end{tabular}
}
\caption{Lean Workbook, $K{=}5$, retrieval-free generator, best-first search.
\textit{Fix}/\textit{Brk} are theorems newly solved/regressed vs.\ the
$R{=}0$ baseline on that split; both splits show \textit{Brk}~=~0 at every
$R$.}
\label{tab:lean_workbook}
\end{table}

Three findings replicate the LeanDojo sweep on this independent benchmark.
First, repair produces a large gain with a weaker, retrieval-free generator:
two repairs take val Pass@1 from 21.4\% to 42.9\%, almost exactly double,
and test Pass@1 from 19.1\% to 31.5\%, a 65\% relative gain. Second, the gain
is strictly additive on both splits -- \textit{Brk}~=~0 at every $R$, so every
baseline proof survives and every new proof is a repair-only unlock, the
cleanest version of the Fix-over-Brk pattern in the paper (the LeanDojo sweep
has small \textit{Brk} at $K{=}5,8$; Table~\ref{tab:main_results}). Third, the
front-loading result replicates quantitatively: the first repair ($R{=}0\to1$)
already captures 18 of the 21 val theorems ($85.7\%$) and 10 of the 11 test
theorems ($90.9\%$) that two repairs ever unlock -- squarely inside the
84--96\% range measured on the LeanDojo sweep. The cost also matches the
earlier characterization: repair is slower than sampling, with a mean latency
of 44.9\,s per repaired proof at val $R{=}2$ (89.5\,s on test, whose
theorems are individually harder), bounded by the shared 600\,s timeout.

\subsection{Qualitative Analysis}

Table~\ref{tab:main_results} and Figure~\ref{fig:repair-saturation} show the
\emph{cumulative} effect of repair, counted from $R{=}0$. Table~\ref{tab:repair_dynamics}
instead isolates the \emph{marginal} cost and gain of each individual repair
attempt---how much wall-clock the step from $R{=}r$ to $R{=}r{+}1$ adds, and how
many additional theorems it buys, as successive differences of
Table~\ref{tab:main_results}'s $T(s)$ and Net columns---and divides the two into
an efficiency figure (theorems gained per added second) so the collapse from the
first repair to later ones is visible directly, rather than requiring the reader
to subtract Table~\ref{tab:main_results}'s rows by hand.

\begin{table}[tbp]
\centering
\setlength{\tabcolsep}{4pt}
{\small
\begin{tabular}{@{}c l r r r@{}}
\toprule
$K$ & Step & $\Delta T$(s) & $\Delta$Net & Eff. \\
\midrule
\multirow{3}{*}{1} & $R{=}0\to1$ &    0.6 & 38 & 63.33 \\
                    & $R{=}1\to2$ &    1.5 &  7 &  4.67 \\
                    & $R{=}2\to3$ &    1.2 &  0 &  0.00 \\
\midrule
\multirow{3}{*}{3} & $R{=}0\to1$ &   29.0 & 57 &  1.97 \\
                    & $R{=}1\to2$ &   14.1 & 10 &  0.71 \\
                    & $R{=}2\to3$ &  $-$0.9 &  1 & $^\dagger$-- \\
\midrule
\multirow{3}{*}{5} & $R{=}0\to1$ &   75.6 & 71 &  0.94 \\
                    & $R{=}1\to2$ &   30.3 &  9 &  0.30 \\
                    & $R{=}2\to3$ &   14.8 &  2 &  0.14 \\
\midrule
\multirow{3}{*}{8} & $R{=}0\to1$ &  166.8 & 77 &  0.46 \\
                    & $R{=}1\to2$ &   82.6 &  2 &  0.02 \\
                    & $R{=}2\to3$ &   35.7 &  1 &  0.03 \\
\bottomrule
\end{tabular}
}
\caption{Marginal cost and gain of each additional repair attempt, by $K$
(successive differences of Table~\ref{tab:main_results}'s $T(s)$ and Net
columns; not cumulative). Eff.~$=\Delta$Net$/\Delta T$, theorems gained per
added second of mean cost. $^\dagger$At $K{=}3$, $R{=}2\to3$, $\Delta T$ is
within measurement noise (Table~\ref{tab:main_results}'s $T(s)$ is nearly flat
there), so efficiency is not meaningfully defined.}
\label{tab:repair_dynamics}
\end{table}

The $R{=}0\to1$ step is the only one where $\Delta$Net is large at every $K$;
every later step adds a small, and at $K{=}8$ nearly negligible, number of solves
at rising cost. The Eff.\ column makes this concrete: at $K{=}8$ the first
repair buys roughly 0.46 more solved theorems per added second of mean cost,
the second and third only 0.02--0.03 -- a more than fifteen-fold collapse in
marginal efficiency, at every $K$ we tested.

\begin{table}[tbp]
\centering
\setlength{\tabcolsep}{5pt}
{\small
\begin{tabular}{@{}l r r r@{}}
\toprule
\textbf{Failure category} & \textbf{Fail} & \textbf{Fix} & \textbf{Rate} \\
\midrule
No progress / simp    & 10{,}110 & 1{,}476 & 14.6\% \\
Apply/exact failed    &  1{,}600 &    212  & 13.2\% \\
Rfl failed            &  8{,}847 &    976  & 11.0\% \\
Unknown identifier    &  7{,}164 &    386  &  5.4\% \\
Rewrite failed        & 16{,}141 &    760  &  4.7\% \\
Typeclass / synthesis &  4{,}712 &     95  &  2.0\% \\
Syntax / parse        &  3{,}731 &     76  &  2.0\% \\
Type mismatch         & 17{,}839 &    328  &  1.8\% \\
Unsolved / timeout    &  2{,}391 &     20  &  0.8\% \\
Other                 & 17{,}099 & 1{,}214 &  7.1\% \\
\midrule
Total                 & 89{,}634 & 5{,}543 &  6.2\% \\
\bottomrule
\end{tabular}
}
\caption{Repair fix rate by failure category, for the $K{=}8$, $R{=}1$ run.
\textit{Fail} counts rejected generator tactics of each type; \textit{Fix}
counts those a single repair turned into a valid successor proof state;
\textit{Rate}~=~Fix/Fail. Ordered by rate; computed from the run logs.}
\label{tab:failure_taxonomy}
\end{table}

Table~\ref{tab:failure_taxonomy} suggests why. Of the 89{,}634 rejected tactics
in the $K{=}8$, $R{=}1$ run, a single repair recovers only 6.2\% into a valid
successor state -- a narrow slice (note the unit: rejected \emph{tactics}, not
theorems as in Table~\ref{tab:main_results}). Repair succeeds mainly on shallow,
local tactic-execution failures --- no progress, a mis-applied
\texttt{apply}/\texttt{exact}, or a non-closing \texttt{rfl} --- where a small
edit suffices. It rarely helps structural or semantic failures such as type
mismatches, typeclass synthesis, or timeouts, which need a different proof
strategy, not a localized fix; missing premises (\texttt{unknown identifier})
fall in between. Because only a small, mostly-shallow fraction of failures is
repairable in principle, one repair exhausts most of the available gain,
consistent with the elbow at $R{=}1$ holding across every $K$; the exact rates
will shift with stronger generators, repair models, and theorem distributions,
but the recipe transfers: weigh unlocks, regressions, and runtime against pure
sampling, and budget at most one or two repairs.

\subsection{Limitations}
The main cost of repair is wall-clock time. Every repair invokes a separate
4-billion-parameter repair model -- heavier than the \texttt{byt5-small}
generator -- so a repair step costs more than a fresh sample, and the overhead
compounds with breadth: enabling repair raises the mean per-theorem time from
200.5 to 485.7\,s at $K{=}8$ (a $2.4\times$ increase), with the first repair
alone adding 166.8\,s. Its coverage advantage does survive a
matched per-theorem budget (Figure~\ref{fig:coverage-time}), but at a very tight
budget the added per-state overhead can delay solves past the deadline, and the
higher latency counts against repair wherever per-theorem time, not solve rate,
is the binding constraint. The overhead also interacts with the
fixed 600\,s timeout: the baseline regressions we observe (Brk) appear only at
the slowest configurations ($K{=}5,8$), where repair's added latency exhausts
the budget on theorems the faster baseline solved in time.

\section{Conclusion}
This paper shows that a bounded, training-free repair step, added inside proof
search, improves on the \emph{generation-first} paradigm that LeanDojo and
ReProver-style provers embody: sample many candidate tactics, keep whichever
the compiler accepts, and discard the rest. Prior work treats repair as a
heavyweight standalone pipeline and has not isolated this in-search question;
we do, and at every candidate budget $K$ tested it recovers proofs pure
sampling never finds -- up to 82 net theorems and a 4.1-percentage-point
Pass@1 gain -- while remaining cheap and well-scoped: a single attempt
captures 84--96\% of the total benefit, the advantage is largest when
sampling is constrained, and it comes almost entirely from a narrow,
identifiable slice of repairable failures. The practical recipe is therefore
simple: add one repair step to a generation-first prover, lean on it hardest
when sampling is scarce, and treat further repairs, especially at large
branching factors, as rarely worth the added wall-clock.

\FloatBarrier
\bibliography{references}

\begin{thebibliography}{20}
\providecommand{\natexlab}[1]{#1}

\bibitem[{Achim et~al.(2025)Achim, Best, Bietti, Der, F{\'e}d{\'e}rico, Gukov, Halpern-Leistner, Henningsgard, Kudryashov, Meiburg, Michelsen, Patterson, Rodriguez, Scharff, Shanker, Sicca, Sowrirajan, Swope, Tamas, Tenev, Thomm, Williams, and Wu}]{achim_aristotle_2025}
Achim, T.; Best, A.; Bietti, A.; Der, K.; F{\'e}d{\'e}rico, M.; Gukov, S.; Halpern-Leistner, D.; Henningsgard, K.; Kudryashov, Y.; Meiburg, A.; Michelsen, M.; Patterson, R.; Rodriguez, E.; Scharff, L.; Shanker, V.; Sicca, V.; Sowrirajan, H.; Swope, A.; Tamas, M.; Tenev, V.; Thomm, J.; Williams, H.; and Wu, L. 2025.
\newblock {{Aristotle}}: {{IMO-level Automated Theorem Proving}}.
\newblock arXiv:2510.01346.

\bibitem[{Brown et~al.(2024)Brown, Juravsky, Ehrlich, Clark, Le, R{\'e}, and Mirhoseini}]{brown_large_2024}
Brown, B.; Juravsky, J.; Ehrlich, R.; Clark, R.; Le, Q.~V.; R{\'e}, C.; and Mirhoseini, A. 2024.
\newblock Large {{Language Monkeys}}: {{Scaling Inference Compute}} with {{Repeated Sampling}}.
\newblock arXiv:2407.21787.

\bibitem[{Chen et~al.(2025)Chen, Gu, Huang, Huang, Jiang, Jie, Jin, Jin, Li, Ma, Ren, Shen, Shi, Sun, Sun, Wang, Wang, Wang, Wei, Wei, Wu, Wu, Xia, Xin, Yang, Ying, Yuan, Yuan, Zhan, Zhang, Zhang, Zhang, Zhao, Zhao, Zhou, and Zhu}]{chen_seedprover_2025}
Chen, L.; Gu, J.; Huang, L.; Huang, W.; Jiang, Z.; Jie, A.; Jin, X.; Jin, X.; Li, C.; Ma, K.; Ren, C.; Shen, J.; Shi, W.; Sun, T.; Sun, H.; Wang, J.; Wang, S.; Wang, Z.; Wei, C.; Wei, S.; Wu, Y.; Wu, Y.; Xia, Y.; Xin, H.; Yang, F.; Ying, H.; Yuan, H.; Yuan, Z.; Zhan, T.; Zhang, C.; Zhang, Y.; Zhang, G.; Zhao, T.; Zhao, J.; Zhou, Y.; and Zhu, T.~H. 2025.
\newblock {{Seed-Prover}}: {{Deep and Broad Reasoning for Automated Theorem Proving}}.
\newblock arXiv:2507.23726.

\bibitem[{Chen et~al.(2021)Chen, Tworek, Jun, Yuan, Pinto, Kaplan, Edwards, Burda, Joseph, Brockman, Ray, Puri, Krueger, Petrov, Khlaaf, Sastry, Mishkin, Chan, Gray, Ryder, Pavlov, Power, Kaiser, Bavarian, Winter, Tillet, Such, Cummings, Plappert, Chantzis, Barnes, Herbert-Voss, Guss, Nichol, Paino, Tezak, Tang, Babuschkin, Balaji, Jain, Saunders, Hesse, Carr, Leike, Achiam, Misra, Morikawa, Radford, Knight, Brundage, Murati, Mayer, Welinder, McGrew, Amodei, McCandlish, Sutskever, and Zaremba}]{chen_evaluating_2021}
Chen, M.; Tworek, J.; Jun, H.; Yuan, Q.; Pinto, H. P. d.~O.; Kaplan, J.; Edwards, H.; Burda, Y.; Joseph, N.; Brockman, G.; Ray, A.; Puri, R.; Krueger, G.; Petrov, M.; Khlaaf, H.; Sastry, G.; Mishkin, P.; Chan, B.; Gray, S.; Ryder, N.; Pavlov, M.; Power, A.; Kaiser, L.; Bavarian, M.; Winter, C.; Tillet, P.; Such, F.~P.; Cummings, D.; Plappert, M.; Chantzis, F.; Barnes, E.; Herbert-Voss, A.; Guss, W.~H.; Nichol, A.; Paino, A.; Tezak, N.; Tang, J.; Babuschkin, I.; Balaji, S.; Jain, S.; Saunders, W.; Hesse, C.; Carr, A.~N.; Leike, J.; Achiam, J.; Misra, V.; Morikawa, E.; Radford, A.; Knight, M.; Brundage, M.; Murati, M.; Mayer, K.; Welinder, P.; McGrew, B.; Amodei, D.; McCandlish, S.; Sutskever, I.; and Zaremba, W. 2021.
\newblock Evaluating Large Language Models Trained on Code.
\newblock arXiv:2107.03374.

\bibitem[{Hubert et~al.(2025)Hubert, Mehta, Sartran, Horv{\'a}th, {\v Z}u{\v z}i{\'c}, Wieser, Huang, Schrittwieser, Schroecker, Masoom, Bertolli, Zahavy, Mandhane, Yung, Beloshapka, Ibarz, Veeriah, Yu, Nash, Lezeau, Mercuri, S{\"o}nne, Mehta, Davies, Zheng, Pedregosa, Li, Von~Glehn, Rowland, Albanie, Velingker, Schmitt, Lockhart, Hughes, Michalewski, Sonnerat, Hassabis, Kohli, and Silver}]{Nature/HubertRLMGEAJYHOTAJIBVLOPSCBADFYIMSASEEHNDP25}
Hubert, T.; Mehta, R.; Sartran, L.; Horv{\'a}th, M.~Z.; {\v Z}u{\v z}i{\'c}, G.; Wieser, E.; Huang, A.; Schrittwieser, J.; Schroecker, Y.; Masoom, H.; Bertolli, O.; Zahavy, T.; Mandhane, A.; Yung, J.; Beloshapka, I.; Ibarz, B.; Veeriah, V.; Yu, L.; Nash, O.; Lezeau, P.; Mercuri, S.; S{\"o}nne, C.; Mehta, B.; Davies, A.; Zheng, D.; Pedregosa, F.; Li, Y.; Von~Glehn, I.; Rowland, M.; Albanie, S.; Velingker, A.; Schmitt, S.; Lockhart, E.; Hughes, E.; Michalewski, H.; Sonnerat, N.; Hassabis, D.; Kohli, P.; and Silver, D. 2025.
\newblock Olympiad-Level Formal Mathematical Reasoning with Reinforcement Learning.
\newblock \emph{Nature}.

\bibitem[{Liu et~al.(2025)Liu, Sun, Li, and Yao}]{liu_proofaug_2025}
Liu, H.; Sun, J.; Li, Z.; and Yao, A.~C. 2025.
\newblock {{ProofAug}}: {{Efficient Neural Theorem Proving}} via {{Fine-grained Proof Structure Analysis}}.
\newblock arXiv:2501.18310.

\bibitem[{Lu, Delaware, and Zhang(2024)}]{lu_proof_2024}
Lu, M.; Delaware, B.; and Zhang, T. 2024.
\newblock Proof {{Automation}} with {{Large Language Models}}.
\newblock arXiv:2409.14274.

\bibitem[{M{\"u}ndler et~al.(2025)M{\"u}ndler, He, Wang, Sen, Song, and Vechev}]{Proc.ACMProgram.Lang./MundlerJHKD25}
M{\"u}ndler, N.; He, J.; Wang, H.; Sen, K.; Song, D.; and Vechev, M. 2025.
\newblock Type-{{Constrained Code Generation}} with {{Language Models}}.
\newblock \emph{Reproduction Package for article "Type-Constrained Code Generation with Language Models"}, 9(PLDI): 171:601--171:626.

\bibitem[{Nagy et~al.(2026)Nagy, Zhou, Polikarpova, and D'Antoni}]{Proc.ACMProgram.Lang./NagyTN26a}
Nagy, S.; Zhou, T.; Polikarpova, N.; and D'Antoni, L. 2026.
\newblock {{ChopChop}}: {{A Programmable Framework}} for {{Semantically Constraining}} the {{Output}} of {{Language Models}}.
\newblock \emph{ChopChop Artifact}, 10(POPL): 66:1905--66:1932.

\bibitem[{Ospanov, Farnia, and Yousefzadeh(2025)}]{ospanov_apollo_2025}
Ospanov, A.; Farnia, F.; and Yousefzadeh, R. 2025.
\newblock {{APOLLO}}: {{Automated LLM}} and {{Lean Collaboration}} for {{Advanced Formal Reasoning}}.
\newblock arXiv:2505.05758.

\bibitem[{Requena et~al.(2026)Requena, Letson, Nowakowski, Beltran-Ferreiro, and Sarra}]{requena_minimal_2026}
Requena, B.; Letson, A.; Nowakowski, K.; Beltran-Ferreiro, I.; and Sarra, L. 2026.
\newblock A {{Minimal Agent}} for {{Automated Theorem Proving}}.
\newblock Accepted for publication at ICML 2026, arXiv:2602.24273.

\bibitem[{Shen et~al.(2025)Shen, Huang, Yang, Wang, Gao, Xu, Jiang, He, Yang, Sun, Ju, Wu, Dai, and Dong}]{shen_real-prover_2025}
Shen, Z.; Huang, N.; Yang, F.; Wang, Y.; Gao, G.; Xu, T.; Jiang, J.; He, W.; Yang, P.; Sun, M.; Ju, H.; Wu, P.; Dai, B.; and Dong, B. 2025.
\newblock {{REAL-Prover}}: {{Retrieval Augmented Lean Prover}} for {{Mathematical Reasoning}}.
\newblock arXiv:2505.20613.

\bibitem[{Song, Yang, and Anandkumar(2025)}]{song_lean_2025}
Song, P.; Yang, K.; and Anandkumar, A. 2025.
\newblock Lean {{Copilot}}: {{Large Language Models}} as {{Copilots}} for {{Theorem Proving}} in {{Lean}}.
\newblock arXiv:2404.12534.

\bibitem[{Wang et~al.(2026)Wang, Chess, Lee, Ge, Mallavarapu, and Ilin}]{wang_learning_2026}
Wang, E.; Chess, S.; Lee, D.; Ge, S.; Mallavarapu, A.; and Ilin, V. 2026.
\newblock Learning to {{Repair Lean Proofs}} from {{Compiler Feedback}}.
\newblock arXiv:2602.02990.

\bibitem[{Wang et~al.(2024)Wang, Xin, Liu, Li, Huang, Lu, Yang, Tang, Yin, Li, and Liang}]{Proc.38thInt.Conf.NeuralInf.Process.Syst./WangHZWYJZJJZ24}
Wang, H.; Xin, H.; Liu, Z.; Li, W.; Huang, Y.; Lu, J.; Yang, Z.; Tang, J.; Yin, J.; Li, Z.; and Liang, X. 2024.
\newblock Proving Theorems Recursively.
\newblock In \emph{Proceedings of the 38th {{International Conference}} on {{Neural Information Processing Systems}}}, volume~37 of \emph{{{NIPS}} '24}, 86720--86748. Red Hook, NY, USA: Curran Associates Inc.
\newblock ISBN 979-8-3313-1438-5.

\bibitem[{Wu et~al.(2025)Wu, Huang, Zhou, Ying, Yuan, Zhang, Lin, and Chen}]{wu_internlm25_2025}
Wu, Z.; Huang, S.; Zhou, Z.; Ying, H.; Yuan, Z.; Zhang, W.; Lin, D.; and Chen, K. 2025.
\newblock {{InternLM2}}.5-{{StepProver}}: {{Advancing Automated Theorem Proving}} via {{Critic-Guided Search}}.
\newblock arXiv:2410.15700.

\bibitem[{Yang et~al.(2023)Yang, Swope, Gu, Chalamala, Song, Yu, Godil, Prenger, and Anandkumar}]{NeurIPS2023/YangAARPSSR23}
Yang, K.; Swope, A.~M.; Gu, A.; Chalamala, R.; Song, P.; Yu, S.; Godil, S.; Prenger, R.; and Anandkumar, A. 2023.
\newblock {{LeanDojo}}: {{Theorem Proving}} with {{Retrieval-Augmented Language Models}}.
\newblock \emph{NeurIPS 2023}.

\bibitem[{Yang et~al.(2025)Yang, Zhou, Wang, Li, Wei, Jin, Li, and Li}]{yang_carts_2025}
Yang, X.-W.; Zhou, Z.; Wang, H.; Li, A.; Wei, W.-D.; Jin, H.; Li, Z.; and Li, Y.-F. 2025.
\newblock {{CARTS}}: {{Advancing Neural Theorem Proving}} with {{Diversified Tactic Calibration}} and {{Bias-Resistant Tree Search}}.
\newblock In \emph{The Thirteenth International Conference on Learning Representations}.
\newblock ICLR 2025 poster.

\bibitem[{Yin and Gao(2025)}]{yin_generating_2025}
Yin, D.; and Gao, J. 2025.
\newblock Generating {{Millions Of Lean Theorems With Proofs By Exploring State Transition Graphs}}.
\newblock arXiv:2503.04772.

\bibitem[{Ying et~al.(2024)Ying, Wu, Geng, Yuan, Lin, and Chen}]{ying_lean_2024}
Ying, H.; Wu, Z.; Geng, Y.; Yuan, Z.; Lin, D.; and Chen, K. 2024.
\newblock Lean {{Workbook}}: {{A}} Large-Scale {{Lean}} Problem Set Formalized from Natural Language Math Problems.
\newblock arXiv:2406.03847.

\end{thebibliography}
\end{document}